\documentclass[conference]{IEEEtran}
\IEEEoverridecommandlockouts
\usepackage{cite}
\usepackage{amsmath,amssymb,amsfonts}
\usepackage{algorithmic}
\usepackage{graphicx}
\usepackage{textcomp}
\usepackage{xcolor}
\def\BibTeX{{\rm B\kern-.05em{\sc i\kern-.025em b}\kern-.08em
    T\kern-.1667em\lower.7ex\hbox{E}\kern-.125emX}}
\begin{document}

\title{Expansion of Cyber Attack Data From Unbalanced Datasets Using Generative Techniques\\
%{\footnotesize \textsuperscript{*}Note: Sub-titles are not captured in Xplore and should not be used}
\thanks{\IEEEauthorrefmark{1} authors have same contribution to this work.}
}

\author{

%%%%%%%%%%% ------AUTHORS--------

\IEEEauthorblockN{Ibrahim Yilmaz\IEEEauthorrefmark{1},  
Rahat Masum\IEEEauthorrefmark{1},}
\IEEEauthorblockA{
\textit{Department of Computer Science}\\
\textit{Tennessee Tech University}\\
Cookeville, TN, USA
}
}
\iffalse
%email address or ORCID}
\and
\IEEEauthorblockN{Benjamin A. Blakely}
\IEEEauthorblockA{
\textit{Argonne National Laboratory}\\
Des Moines, Iowa, USA
}
\and
\IEEEauthorblockN{Ambareen Siraj}
\IEEEauthorblockA{
\textit{Department of Computer Science}\\
\textit{Tennessee Tech University}\\
Cookeville, TN, USA
}
%email address or ORCID}

  Ibrahim Yilmaz\IEEEauthorrefmark{1},
  Rahat Masum\IEEEauthorrefmark{1},
  Benjamin A. Blakely\IEEEauthorrefmark{2},
  Ambareen Siraj\IEEEauthorrefmark{1}
 %Mohammad Ashiqur Rahman\IEEEauthorrefmark{3}
  \\
  \IEEEauthorblockA{
    \IEEEauthorrefmark{1}Department of Computer Sience, Tennessee Tech University, Cookeville, TN, USA
  }
   \IEEEauthorblockA{
    \IEEEauthorrefmark{2}Argonne National Labs
  }
  \fi
  %\IEEEauthorblockA{
   % \IEEEauthorrefmark{3}Department of Electrical and Computer Engineering, Florida International University, Miami, FL, USA}

\maketitle

\begin{abstract}
Machine learning techniques help to understand patterns of a dataset to create a defense mechanism against cyber attacks. However, it is difficult to construct a theoretical model due to the imbalances in the dataset for discriminating attacks from the overall dataset. Multilayer Perceptron (MLP) technique will provide improvement in accuracy and increase the performance of detecting the attack and benign data from a balanced dataset. We have worked on the UGR'16 dataset publicly available for this work. Data wrangling has been done due to prepare test set from in the original set. We fed the neural network classifier larger input to the neural network in an increasing manner (i.e. 10000, 50000, 1 million) to see the distribution of features over the accuracy. We have implemented a GAN model that can produce samples of different attack labels (e.g. blacklist, anomaly spam, ssh scan). We have been able to generate as many samples as necessary based on the data sample we have taken from the UGR’16. We have tested the accuracy of our model with the imbalance dataset initially and then with the increasing the attack samples and found improvement of classification performance for the latter.
\end{abstract}


\section{Introduction}
Machine learning approaches should use stationary domain, where train-test data comes from the same origin. However, if the ratio between the class samples is not balanced, the designed classifier may predict wrong over malicious data as well as benign data. Adversarial training is possible to generate inputs and explicitly train the model, which will help cover a wide range of distribution for data samples fed to the classifier model \cite{Attackin87:online}~\cite{WhatisAd63:online}. The publicly available datasets are either not up-to-date or do not preserve the balance between attack vs. non-attack classes. Therefore, available dataset for the IDS sector needs to be balanced with more attack data label. Our approach is to analyze the existing dataset to determine discrepancies and to prepare the dataset for use in the neural network classifier.

One of the goals of using the neural network is to provide/generate more samples of attack data labels so that training of the model may be robust and widely covered in case of intrusion detection in the network traffic flow analysis. If a \textit{generator} can produce inputs representing the attack class, the balance of the training dataset can be ensured and the GAN model will work more accurately in distinguishing actual attack data. The \textit{discriminator} function built using as a Deep Neural Network (MLP) with GAN, will quantify the scalability and degree of balancing training dataset. To summarize our main contributions:
%Further analysis will help to acquire knowledge of cyber-attack scenario more closely.

\begin{itemize}
    \item We have studied different available datasets and found the best-suited one as UGR'16 to analyze the problem and design solution methods.
    \item We have used different big data analytical techniques to handle the large dataset to prepare the input data for our proposed model.
    \item We have prepared the features with different data types by converting those into the required data design for the training model.
    \item To solve the unbalance dataset problem, we have introduced the Generative Adversarial Network (GAN) technology to produce necessary attack samples from the UGR'16 dataset. These new samples have improved the training of the data distribution providing balance ratio to the non-attack samples and increased the model performance in detecting attack scenarios.
\end{itemize}
The rest of this paper is organized as: Section~\ref{sec:background} provides an overview of the existing approaches to the unbalanced dataset analysis. In Section~\ref{sec:solution} we briefly discuss the methods and approaches for solving the problem and Section~\ref{sec:implementation} illustrates the implementation model. Section V contains our evaluation of the proposed solution from different metrics, and Section~\ref{sec:conclusion} concludes this work with future directions.

\section{Literature Review}
\label{sec:background}
Machine learning results are usually accurate when training data is selected specifically and the results are closely revisited. For analysis of classifier performance, false positives, false negatives are determined by machine learning.

%%%%%%%%%%%%%%%%%%%
Douzas et al. ~\cite{douzas2018effective} provide a conditional version of Generative Adversarial Networks (cGAN) used to approximate the true data distribution and generate data for the minority class of various imbalanced datasets. Their approaches are based on local information, rather than overall minority class distribution.

A Generative Adversarial Network (GAN) is a feedback neural network. GAN enables the way of learning deep representation where training data do not need to be annotated extensively. It involves the backpropagation of signals through a neural network to process and derive the class labels~\cite{kurakin2018ensemble}~\cite{kurakin2016adversarial}. GAN can be a useful technique to represent image synthesis, semantic image editing, and super-resolution. Kurakin et al. provided an overview of GANs about the signal processing community as well. In~\cite{radford2015unsupervised}, an evaluation of the architectural topology of Convolutional GANs has been done that is a stable approach to train settings. The authors show that the deep convolutional adversarial pair can learn partial objects from scenes for \textit{generator} and \textit{discriminator}.

Liu et al.~\cite{liu2008exploratory} proposed the undersampling method to deal with an imbalanced dataset. According to this method, a subset of the majority class was selected and the remaining subsets were ignored. In this manner, the dataset became more balanced and the training of the model was faster. They chose several subsets and trained them independently with machine learning models. They then combined the output of machine learning models.

Chawla et al.~\cite{chawla2002smote} addressed the imbalanced dataset issue and introduced the SMOTE (Synthetic Minority Oversampling Technique) model technique to handle this issue. SMOTE is a statistical technique used to produce synthetic samples for a minority class from existing minority classes in a balanced way without changing the number of majority samples. According to this technique, when input data is plotted into feature space for each minority sample, new data, which is similiar to minority class samples in the feature space, is generated. By generating more synthetic samples, the dataset becomes balanced artificially.

Mariani et al. ~\cite{mariani2018bagan}proposed a balancing generative adversarial network (BAGAN) as an augmentation tool to overcome imbalanced image dataset problems. The authors mentioned that the accuracy of image classification techniques is poor with the imbalanced dataset. Their proposed tool managed to produce many different synthetic images to allow for dataset balance, and their results showed that BAGAN outperformed other generative adversarial networks. From the proposed approach the authors train all images of majority and minority classes and then generate images for minority classes. 
Salehinejad et al.~\cite{salehinejad2018generalization} offered a deep convolutional generative adversarial network (DCGAN) which augmented chest X-ray images in order to turn imbalanced into balance dataset. ~\cite{baza2019b,baza2018blockchain,parksmarnet,baza2019blockchain,parkccnc,Lightride,Andrew} 

From the current literature study, there is a lack in balance with datasets available to classify benign and attack data labels. The available dataset accessible for the research study will need to be analyzed depending on the adversarial scenarios about computational time for correct classification. Possible metrics to understand the feature distribution for classifying training dataset efficiently needs to be investigated. This approach will mitigate the problem of imbalance in the dataset.

\section{Solution Approaches}
\label{sec:solution}

%%%%%%%%%%%%%%%%%%%%%%%%%
\begin{figure}[!h]
\includegraphics[width=0.5\textwidth]{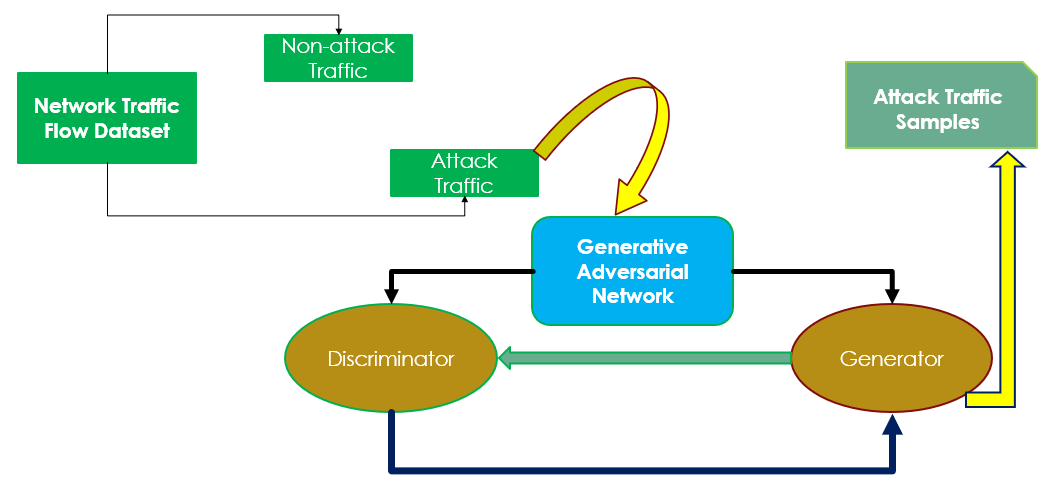}
%\centerline{\includegraphics{work flow.png}}
\caption{Solution Design}
\label{fig:flow}
\end{figure}

%%%%%%%%%%%%%%%%%%%%%%%%%%
Machine learning algorithms are a data-driven approach so that the performance of the model is highly related to data. According to a recent survey, machine learning techniques are vulnerable to data imbalance problems where the number of samples of a  class is greatly outnumbered by the number of the other class sample for binary classification. Addressing this problem presents considerable challenges \cite{rout2018handling}. The UGR’16 dataset is one of the examples of the imbalance dataset, where machine learning techniques are used to find the small number of attack instances. In such circumstances, it is difficult to classify samples as attack or non-attack using traditional machine learning algorithms, since state of the art machine learning models cannot learn the characteristic behavior of the minority attack class. Models are easily biased to the majority class.  This causes the overfitting of the models and consequently, trivial classifiers that predict solely the majority class are built.  As long as the target solution is to maximize the accuracy of the model, the accuracy of the model trained with an imbalanced dataset is highly acceptable. However, if the minority class samples are more crucial, for example, detecting attacks data, then the existence of the dataset should be modified by making it balanced in order to detect attack samples better. Our goal is to create a balanced dataset so that we can investigate through network packet flow for better intrusion detection and analysis. \cite{baza2019detecting,omar1,omar2}

To tackle the imbalanced dataset problem, we propose a GAN methodology to generate necessary synthetic attack samples in order to balance the dataset. Figure~\ref{fig:flow} illustrates the solution model from the data preprocessing to providing balance to the dataset approaches. Our solution design includes two neural network classifiers, labeled \textit{discriminator} (D) and \textit{generator} (G). This was introduced by Goodfellow et al. in 2014 \cite{goodfellow2014generative} by inspiring game theory. These two neural networks-based models are in the competition. GAN is able to produce completely new data similiar to the training data based on the probability distribution model.

The G classifier makes an effort to capture the data distribution of real data to generate synthetic attack samples. In the beginning, G produces a certain degree of random noise and sends it to D. D takes real data and synthetic data which is generated by G as an input and distinguishes them. With fine tuning of parameters of G, we come to see that the real data distribution and the fake data distribution appear to be highly similar to one another. At this point, the fake data produced by G cannot be distinguished by D and thus new synthetic data is created. The mathematical definition of the GAN model is given below~\cite{porikli2018deep}.

\begin{align*}
\min_{G}\max_{D}V(G,D) & = E_{p_{{real\_data}}(x)}log D(x)\\
                       &+E_{p_{{fake\_data}}(x)}log(1-D(x))
\end{align*}

where V(G,D) is a value function that depend on both \textit{discriminator} and \textit{generator}. It is used to calculate cost function. $P_{real\_data}$(x) and $P_{fake\_data}$(x) are the probability of real and the probability of fake data, respectively. $E_{p_{{real\_data}}(x)}$ is an error function for real data and $E_{p_{{fake\_data}}(x)}$ is an error function for fake data which is generated by the \textit{generator}

 Our approach is to find the neural network optimal parameter so that investigation over GAN can give insights about possible cyberattack scenarios. To accomplish this goal, the \textit{generator} can be optimum when $P_{real\_data}$(x) is equal to $P_{fake\_data}$(x) where the optimum \textit{discriminator} predicts all samples as 0.5 probability. To put it in another perspective, the \textit{generator} reaches its optimum point when \textit{discriminator} is  unable to figure out differences between real and fake samples.

\section{Implementation}
\label{sec:implementation}
\subsection{Characterization of Methods}
 The UGR’16 dataset collects data weekly and consists of more than one billion data. This dataset has 12 features named as: timestamp, duration of flow, source IP, destination IP, source port, destination port, protocol, flags, forwarding status, type of service, packets exchanged in the flow, and number of bytes. 
 
 Data come from several netflow v9~\cite{NetFlowV87:online} collectors strategically located in the network of a Spanish ISP. From the UGR’16 dataset, we have focused on the ‘TEST’~\cite{macia2018ugr} set for creating the training model, which has labeled results from the previous classification. This dataset is divided between two months, ‘July’ and ‘August’, which contains a total of 6 (six) weeks of data of network traffic flow. Each of the week's dataset contains approximately one billion data (rows in CSV file). For better computation and analysis, we have preprocessed the dataset to divide into smaller subsets. We have prepared a program using python (pandas) to generate 50,000 rows per file to create a smaller CSV file for our designed classifier.
 
\subsubsection{Dataset Analysis}

We implement a binary classification model to detect attack samples. In order to do that, we create seven different datasets. Each dataset consists of non-attack and a particular attack type, which is one of the 'blacklist', 'anomaly-spam', 'anomaly-sshscan', 'dos', 'nerishbotnet', 'scan 11', 'scan 44' attack family. These datasets have a different number of non-attack and attack samples. In Section~\ref{sec:results}, an analysis of these datasets is discussed in detail. 

In the preprocessing phase, we prepare our data to train the model. Some of the features of the UGR'16 dataset have string values (rather than numeric values). Because  machine learning techniques only deal with numeric values, we converted these string values into numeric values such as values of 'source and destination of IP addresses' mapped to unique numeric values. Then, we normalized our data between 0 and 1 by not compromising the differences of values and their extent, as the normalization gives equal chance to all features to impact the model decision. For instance, the 'type of service' and the 'duration of flow' features have values ranging from 0 to 1, whereas 'packets exchanged in the flow' feature have values ranging from 1 to 4090 for the particular dataset created. These big differences in the range of numbers are the reason for the problems when attempting to merge values to function as features of the model. Normalization prevents the problems as mentioned above with new values created to sustain the general distribution as well as the ratio of the data source while holding the given values in the range of all numeric columns in the model. Visual representations of the data histograms according to each feature are demonstrated in Figure 2, Figure 3, Figure 4 ,and Figure 5, after the data normalization is completed for the blacklist attack dataset.

\begin{figure*}
    \includegraphics[width=1\textwidth]{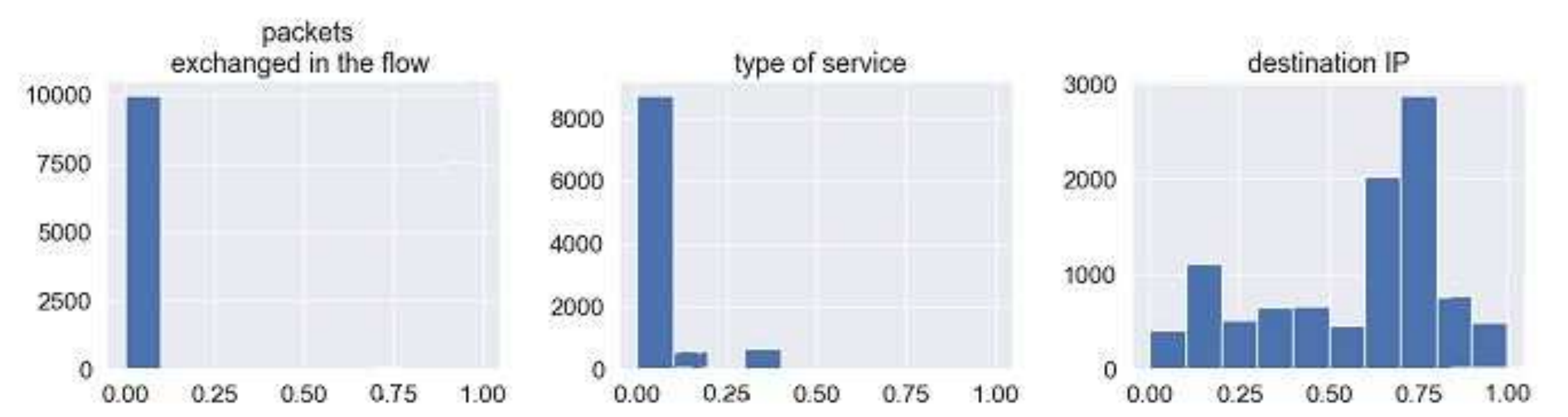}\hfill

    \caption{Data Histogram of the features for (from left to right) Packets Exchanged in the Flow, Type of Service, Destination IP}

\end{figure*}
\begin{figure*}
    \includegraphics[width=1\textwidth]{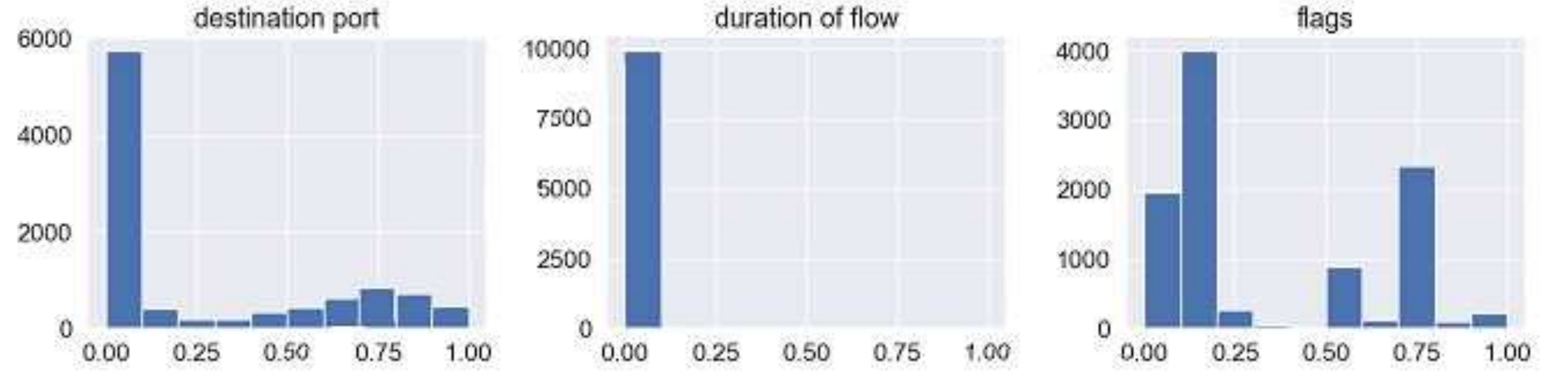}\hfill

    \caption{Data Histogram of the features for (from left to right) Destination Port, Duration of Flow, Flags}

\end{figure*}
\begin{figure*}
    \includegraphics[width=1\textwidth]{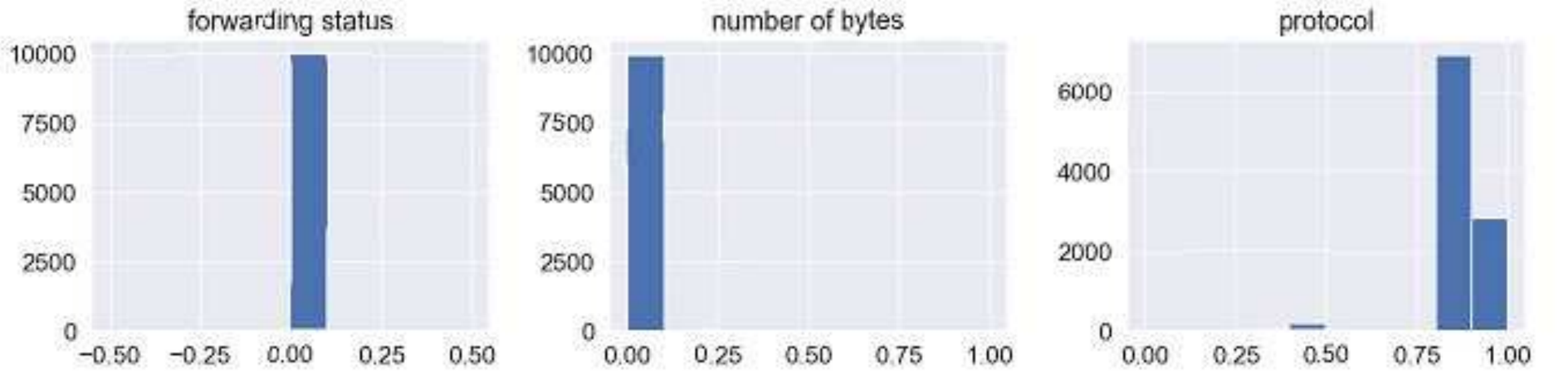}\hfill

    \caption{Data Histogram of the features for (from left to right) Forwarding Status, Number of bytes, Protocol}

\end{figure*}
\begin{figure*}
    \includegraphics[width=0.7\textwidth]{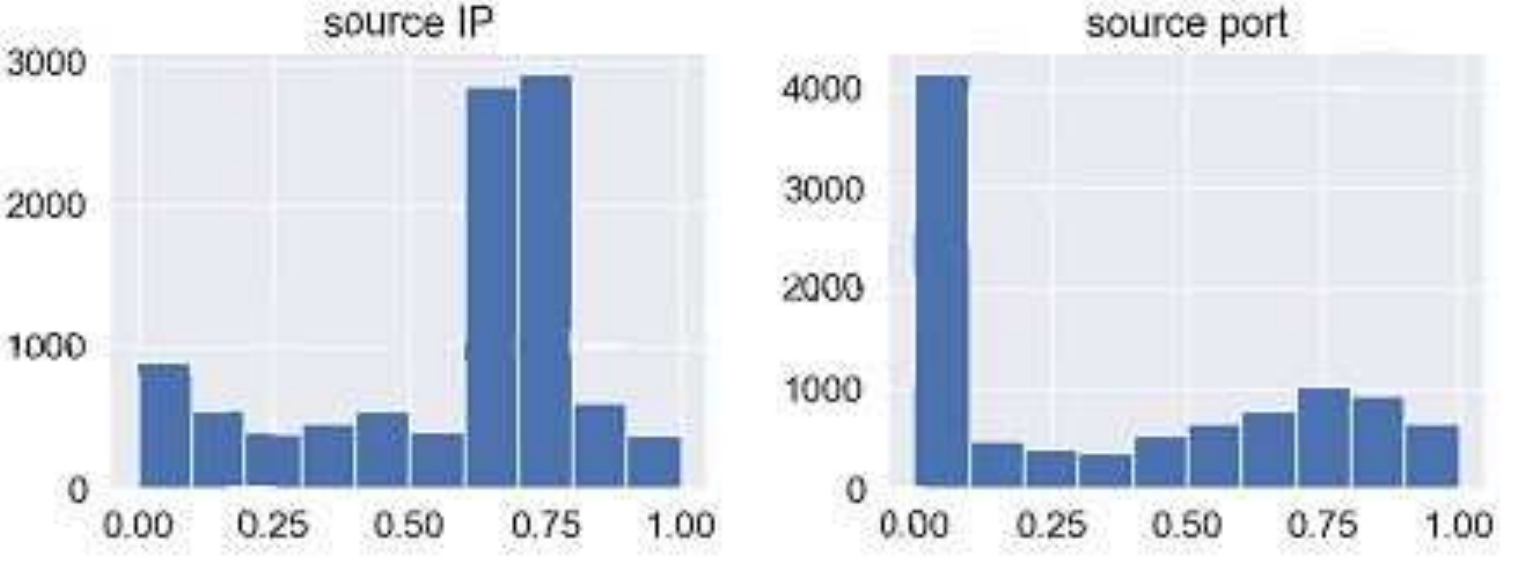}\hfill

    \caption{Data Histogram of the features for (from left to right) Source IP, Source Port}

\end{figure*}

\subsubsection{Generative Adversarial Network Models}

 We first take available UGR'16 dataset to prepare the training model. We divide the whole dataset into smaller parts and create multiple new datasets for each particular attack samples to train the neural network model. In this training, the classifier adjusts its parameter to minimize the error. 
 
 In our analysis, we have used 60\% of the dataset as a training set and 40\% of the dataset as a test set for the classifier. Accuracy and analysis of this classifier are discussed in Section~\ref{sec:results}. To train our models, we use the \textit{Sklearn.Neural\_Network} library and use the \textit{Broyden–Fletcher–Goldfarb–Shanno} (BFGS) optimization solver for adjusting parameters in order to minimize error. We set the number of hidden layers as five (5) and the learning rate as $10^{-6}$.
 
 However, the neural network model cannot learn the characteristic behavior of the attack samples well enough because of the undersampling of the attack samples, regardless of how the neural network model's architecture is built. The neural network performs poorly on this imbalance dataset and always predicts non-attack class for unseen data on the test phase. To tackle this imbalance in the dataset issue, we create a GAN model by implementing \textit{generator} and \textit{discriminator} models using different datasets to complete the classifier. 
 
 We use 5 hidden layers for both the \textit{discriminator} and the \textit{generator} based on a neural network model. Following each hidden layer, the \textit{rectifier liner unit }(ReLU) function is applied. After the output layer, \textit{Sigmoid} function is applied to make it non-linear for both models, which are required to solve complex problems. We set the learning rate by the factor of 0.0025 for the \textit{discriminator} model and by the factor of 0.02 for the \textit{generator} model.  
 
%%%%%%%%%%%%%%%

%%%%%%%%%%%%%%%%%%%%%%%%%%%%

\section{Evaluation}
\label{evaluation}
\subsection{Results}
\label{sec:results}
In this section, we present the findings of our work and evaluate the model performance based on both balanced and imbalanced datasets. To validate the effectiveness of the recommended GAN method, we retrain the model with modified datasets including both real and synthetic data that is generated by GAN. Then we compare results with the previous model's results where the model is trained with imbalanced datasets.

To implement a binary classification model, we construct different sub-datasets derived from the UGR'16 for each malicious family. In order to assess the performance of the model, we use the model measurement such as: accuracy, precision, recall and F1 score ~\cite{precision:online}~\cite{recall:online}~\cite{F1:online}.

Accuracy is the number of true predictions over the number of total predictions of the model.
\begin{equation}
Accuracy = \frac {TP + TN}{FP + FN}
\end{equation}
The recall is the proportion of number of malicious samples was identified correctly as malicious.
\begin{equation}
Recall = \frac {TP}{TP + FN}
\end{equation}
Precision is the proportion of the true predictions of the attack samples over the total prediction of the model as an attack. 
\begin{equation}
Precision = \frac {TP}{TP + FP}
\end{equation}
F1 score is the harmonic average of precision and recall.
\begin{equation}
 F1 \:score =\frac{2*Precision*Recall}{Precision + Recall}
\end{equation}

True Positive (TP) and True Negative (TN) are the numbers that are classified correctly by the model as an attack and a non-attack respectively. False Postive (FP) and False Negative (FN) are the numbers of the misclassified attack and non-attack classes in turn.

Table 1 shows the neural network classifier's performance in terms of accuracy, precision, recall and F1 score with unbalanced datasets while Table 2 demonstrates the same model performance with balanced datasets. Based on our empirical findings, the detection of the malicious activities in the network flow has improved when datasets are made a balance by applying GAN methods.

\begin{table}[hbt!]
\caption{Neural network model performance with unbalanced datasets}
\label{Unbalance}
\resizebox{\columnwidth}{!}{\begin{tabular}{|l|l|l|l|l|}
\hline
\textbf {Dataset Name}&\textbf{Number of}&\textbf{Number of}&\textbf{Statistical}&\textbf{Performance}\\
&\textbf{Attack Samples} &\textbf{Non-attack Samples}&\textbf{Evaluation}&\textbf{Results}\\
\hline
&  & & Accuracy & 99.63\%\\
Blacklist& 60  & 9941 & Precision & 0.0\%\\
&  &  & Recall & 0.0\%\\
&  &  & F1 score & 0.0\%\\
\hline
 &  &  & Accuracy & 99.65\%\\
Anomaly Spam & 70&19.700 & Precision & 50\%\\
&   &  & Recall & 17.86\%\\
&   &  & F1 score & 26.32\%\\
\hline
&  &  & Accuracy & 99.84\%\\
Anomaly Sshscan & 22 & 14.484 & Precision & 0.0\%\\
&   &  & Recall & 0.0\%\\
&   &  & F1 score & 0.0\%\\
\hline
 &  & & Accuracy & 99.97\%\\
Dos& 45 & 101.253 & Precision & 0.0\%\\
 &   &  & Recall & 0.0\%\\
&  &  & F1 score & 0.0\%\\
\hline
&  & & Accuracy & 99.20\%\\
Nerishbotnet& 68 & 10.295 & Precision & 0.0\%\\
 &   &  & Recall & 0.0\%\\
&  &  & F1 score & 0.0\%\\
\hline
&  & & Accuracy & 99.76\%\\
Scan11& 67 & 24.954 & Precision & 66.67\%\\
 &   &  & Recall & 8.0\%\\
&  &  & F1 score & 14.29\%\\
\hline
&  & & Accuracy & 99.24\%\\
Scan44& 71 &11.807  & Precision & 0.0\%\\
 &   &  & Recall & 0.0\%\\
&  &  & F1 score & 0.0\%\\
\hline
\end{tabular}}
\end{table}

\begin{table}[hbt!]
\caption{Neural Network Model Performance with Balanced Datasets}
\label{Unbalance}
\resizebox{\columnwidth}{!}{\begin{tabular}{|l|l|l|l|l|}
\hline
\textbf {Dataset Name}&\textbf{Number of}&\textbf{Number of}&\textbf{Statistical}&\textbf{Performance}\\
&\textbf{Attack Samples} &\textbf{Non-attack Samples}&\textbf{Evaluation}&\textbf{Results}\\
\hline
&  & & Accuracy & 99.69\%\\
Blacklist& 9941  & 9941 & Precision & 99.37\%\\
&  &  & Recall & 100\%\\
&  &  & F1 score & 99.68\%\\
\hline
 &  &  & Accuracy & 99.79\%\\
Anomaly Spam & 19.700&19.700 & Precision & 99.67\%\\
&   &  & Recall & 99.91\%\\
&   &  & F1 score & 99.79\%\\
\hline
&  &  & Accuracy & 99.87\%\\
Anomaly Sshscan & 14.484 & 14.484 & Precision & 99.74\%\\
&   &  & Recall & 100\%\\
&   &  & F1 score & 99.87\%\\
\hline
 &  & & Accuracy & 99.97\%\\
Dos& 101.253 & 101.253 & Precision & 99.97\%\\
 &   &  & Recall & 99.91\%\\
&  &  & F1 score & 99.94\%\\
\hline
&  & & Accuracy & 99.79\%\\
Nerishbotnet& 10.295 & 10.295 & Precision & 99.96\%\\
 &   &  & Recall & 99.60\%\\
&  &  & F1 score & 99.78\%\\
\hline
&  & & Accuracy & 99.80\%\\
Scan11& 24.954 & 24.954 & Precision & 99.75\%\\
 &   &  & Recall & 99.82\%\\
&  &  & F1 score & 99.78\%\\
\hline
&  & & Accuracy & 99.70\%\\
Scan44& 11.807 &11.807  & Precision & 99.58\%\\
 &   &  & Recall & 99.80\%\\
&  &  & F1 score & 99.69\%\\
\hline
\end{tabular}}
\end{table}

Although the accuracy results of the model are satisfactory when the model is trained with unbalanced datasets, this is misleading since the test datasets are imbalanced as well. Even though the model always predicts the majority class for any sample, the calculated accuracy of the values high. Therefore, accuracy is not a good measure when working with an imbalanced dataset.

\begin{figure*}
    \includegraphics[width=1\textwidth]{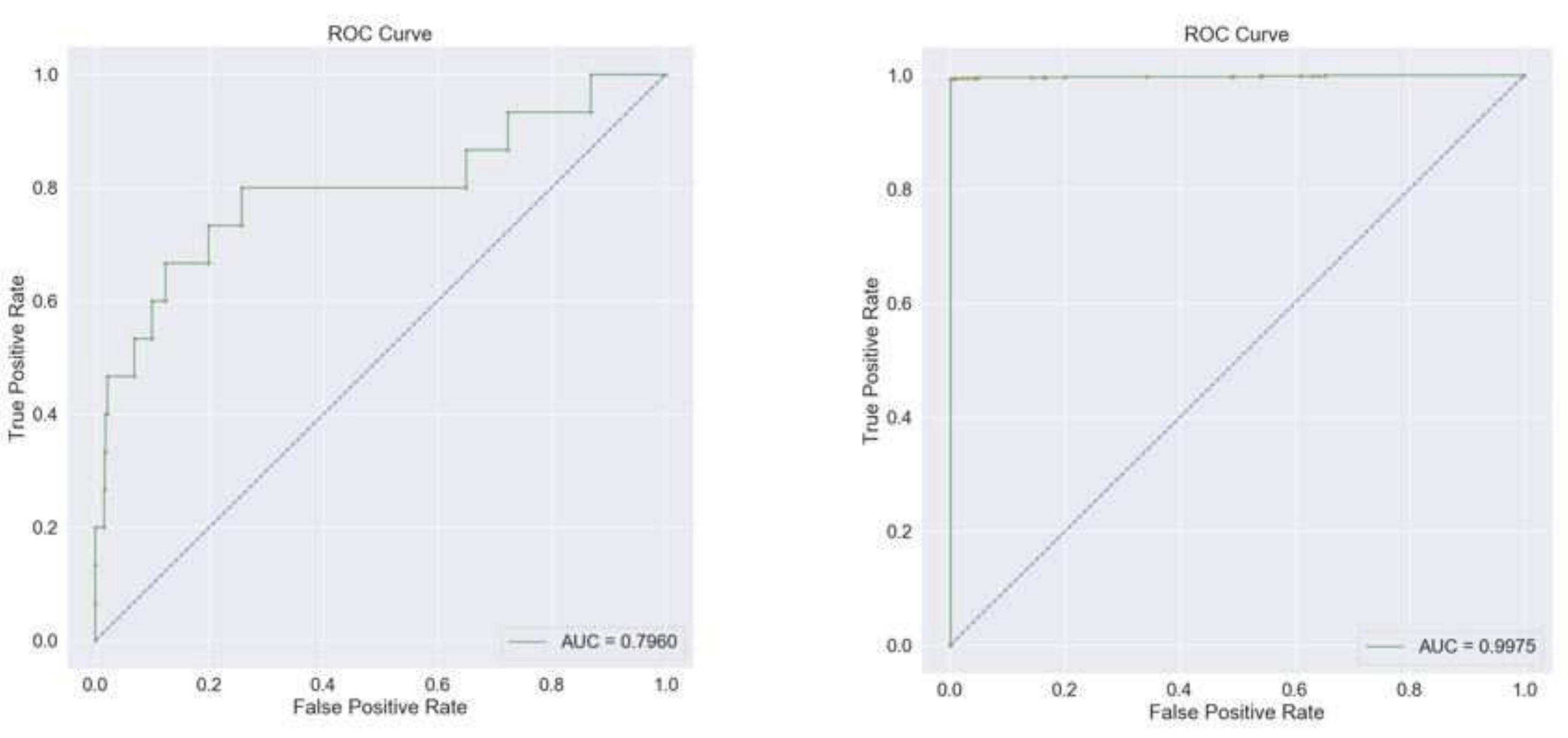}\hfill
    
    \caption{ ROC curve with an unbalanced dataset and ROC curve with a balanced dataset (from left to right) }

\end{figure*}

\begin{figure*}
    \includegraphics[width=1\textwidth]{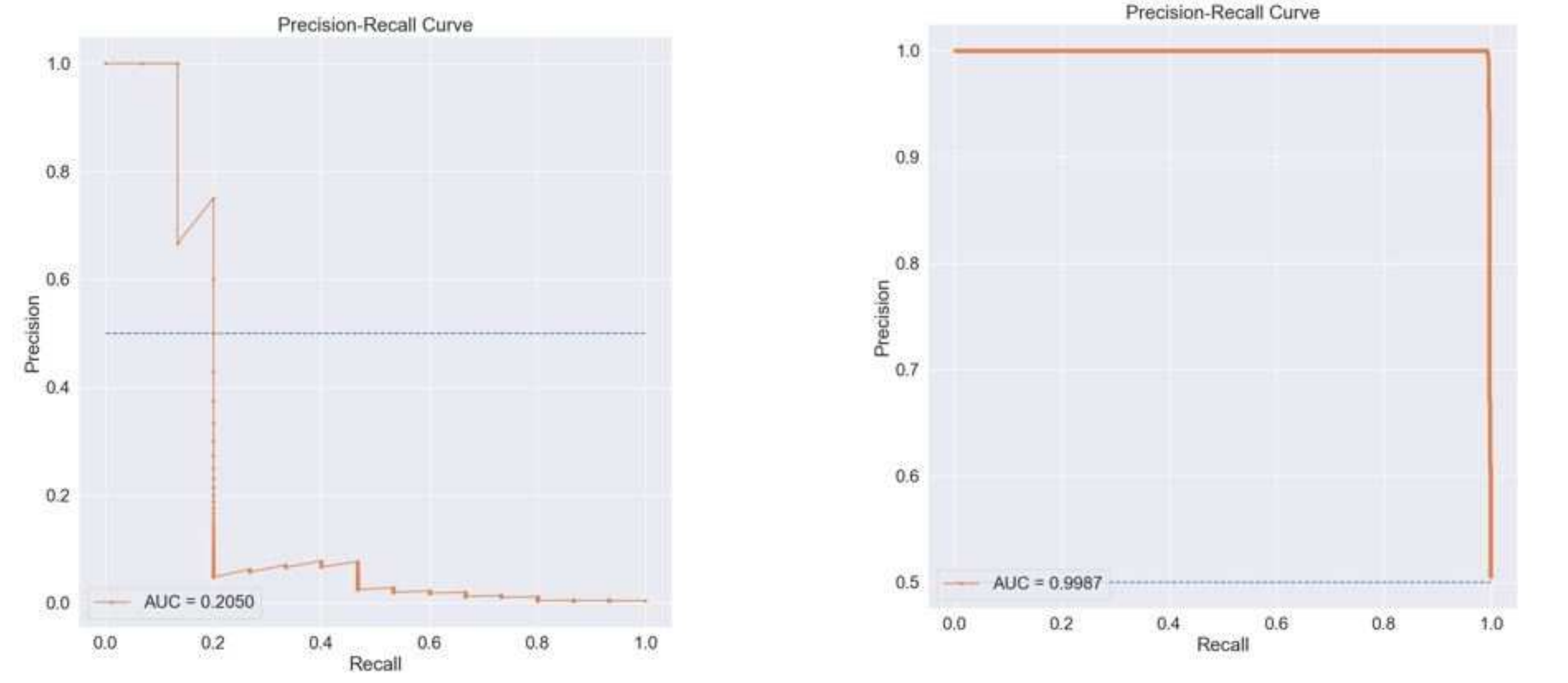}\hfill
    
    \caption{ Precision-Recall curve with an unbalanced dataset and Precision-Recall curve with a balanced dataset (from left to right) }

\end{figure*}

Furthermore, in order to visualize the model performance, we present the Receiver Operating Characteristic (ROC) curve and the Precision-Recall curve for both a balanced and an unbalanced dataset for blacklist attack types. The ROC curve is a way to understand the separation of malicious and benign samples at different threshold settings. An ROC curve is generated using cumulative distribution function. X-axis representing a false positive rate whereas Y-axis representing true positive rate by different threshold values. This threshold values are changed by the importance of false positive rate and true positive rate. If the area under the curve (AUC) value is larger, the performance of the model is better. In Figure 6, it is illustrated that the comparison results of the ROC curve between the unbalanced and the balanced datasets. AUC value in the unbalanced set is around 0.80 and this is caused by the inability of the model to learn the malicious class and overfitting it to the benign class. For this reason, in the test data, the samples are guessed as non-attack by the model. The opposite happens in the balanced dataset, the model can learn from the attack samples and prevent the occurrence of overfitting and this improves the performance of the model. As it can be seen from the graph, the AUC value is very close to 1 and this shows that the performance of the model is almost perfect.

In addition, the Precision-Recall curve is another evaluation method of machine learning models. Precision and recall are inversely related, meaning as one of them increases the other one decreases as this is the way to observe the relations between these two metrics with different threshold values. They are exceptionally helpful with the problems we are having like an unbalanced dataset. Since our datasets are non-attack sample majority, it is extremely important to emphasize the minority. The area under the precision-recall curve also gives very important information about the model performance like the ROC curve does. As it can be seen in Figure 7 AUC value in the unbalanced dataset is close 0.2 and this demonstrates the poor performance of the model, as this value approaches 1 in the balanced dataset and this shows that this model improves the performance by a large margin.

To sum it all up, the model that is trained with these datasets balanced by GAN performs significantly better than the model that was trained with unbalanced datasets.

\subsection{Limitations}
\label{limitations}

ML is very compute intensive and running neural network model over a large/multiple dataset may require high performance GPU and large storage media. For now, we have  worked with our personal computers. Unfortunately, we have limited RAMs in our computers to run all data so we must extract a part of the dataset and work on only a small part of it. Due to the dataset which we have worked on, being unbalanced and not well-formed, we have run some analysis steps to clean the dataset replacing any missing values. We also have mapped some features (e.g. IP address) to integer values to represent mapping in the classifier.

Also, due to the large number of neurons, iterations and acceptable error rate is unpredictable, it mainly depends largely on the neural network architecture and the dataset itself. Furthermore, applying GANs to cyberattack is fairly untested concept; hence, feasibility of the implementation and challenges over the solution was contingent upon multiple factors.

%%%%%%%%%%%%%%%%%%%%%%%%%%%%%%%%%%%%%%%%%%%%%%%%%%

\section{Conclusion}
\label{sec:conclusion}
Since GAN will provide the chance of generating sample data, our approach will be to increase the number of observations belonging to one class than those belonging to the other ones. Hence, we can synthesize data samples of attack class which has a lower number of samples compared to another class (non-attack traffic) to make the dataset more balanced. We have developed a GAN model that will create adversarial samples representing attack classes of the training dataset so that further progress can be made over the improvement of classifier performance mitigating the imbalances on the dataset.

As a future direction, we would like to produce synthetic data using different adversarial techniques such as Saliency Map or Gradient approaches in order to increase the performance of the model and compare results with one which is produced by GAN.

\bibliographystyle{unsrt}
\bibliography{TT}

\begin{thebibliography}{10}
\bibitem{WhatisAd63:online}
What is adversarial machine learning | fresh business thinking.
\end{thebibliography}

\end{document}